\documentclass[conference]{IEEEtran}
\IEEEoverridecommandlockouts
\usepackage{cite}
\usepackage{amsmath,amssymb,amsfonts}
\usepackage{algorithmic}
\usepackage{graphicx}
\usepackage{textcomp}
\usepackage{xcolor}
\usepackage{flushend}
\usepackage{makecell}
\usepackage{booktabs}
\usepackage{dsfont}
\usepackage{caption}
\usepackage{subcaption}
\def\BibTeX{{\rm B\kern-.05em{\sc i\kern-.025em b}\kern-.08em
    T\kern-.1667em\lower.7ex\hbox{E}\kern-.125emX}}
\begin{document}

\title{\textbf{GAANet}: \textbf{G}host \textbf{A}uto \textbf{A}nchor \textbf{Network} for Detecting Varying Size Drones in Dark}
\author{\IEEEauthorblockN{Misha Urooj Khan${^*}$, Maham Misbah${^*}$, Zeeshan Kaleem${^*}$, Yansha Deng${^\ddag}$, Abbas Jamalipour${^ \dag}$}}
\maketitle
\begin{abstract}
The usage of drones has tremendously increased in different sectors spanning from military to industrial applications. Despite all the benefits they offer, their misuse can lead to mishaps, and tackling them becomes more challenging particularly at night due to their small size and low visibility conditions. To overcome those limitations and improve the detection accuracy at night, we propose an object detector called Ghost Auto Anchor Network (GAANet) for infrared (IR) images. The detector uses a YOLOv5 core to address challenges in object detection for IR images, such as poor accuracy and a high false alarm rate caused by extended altitudes, poor lighting, and low image resolution. To improve performance, we implemented auto anchor calculation, modified the conventional convolution block to ghost-convolution, adjusted the input channel size, and used the AdamW optimizer. To enhance the precision of multiscale tiny object recognition, we also introduced an additional extra-small object feature extractor and detector. Experimental results in a custom IR dataset with multiple classes (birds, drones, planes, and helicopters) demonstrate that GAANet shows improvement compared to state-of-the-art detectors. In comparison to GhostNet-YOLOv5, GAANet has higher overall mean average precision (mAP@50), recall, and precision around 2.5\%, 2.3\%, and 1.4\%, respectively. The dataset and code for this paper are available as open source at https://github.com/ZeeshanKaleem/GhostAutoAnchorNet.
\end{abstract}


\section{Introduction}
Unmanned aerial vehicles (UAVs) are widely adopted in remote sensing and advanced surveillance applications due to the growth in drone-based applications. According to industry insights, the global drone market is expected to reach \$48 billion by 2026 \cite{1}. Because of their flexibility and mobility, drones are widely considered in many daily and industrial applications, and their capabilities are further enhanced when equipped with advanced artificial intelligence (AI) techniques. This advancement and its increasingly widespread use have raised serious concerns about the security of public places, as we have seen several instances where drones have caused damage to infrastructure \cite{2}\cite{3}. Therefore, effective detection systems are necessary for protection against malicious activities \cite{4}. Advanced object detection and tracking systems are preferred over traditional object identification methods due to their less accurate target detection and high false alarm rate. Object identification is increasingly adopted for drone detection, but many schemes fail because of the drone's small size, high flight altitude, and fast speed. These issues are addressed by UAV detection systems integrated with deep learning algorithms. Object detection using computer vision and deep learning, such as regions with convolutional neural networks (RCNN), Faster RCNN \cite{r4}, and Mask-RCNN, utilize two-stage detection methods with improved detection results. But they are unsuited for efficient and accurate recognition of tiny fast-moving objects like UAVs vs. birds, planes, or helicopters. You Only Look Once (YOLO) \cite{r5} and the single shot multibox detector (SSD) \cite{r6} are two more methods that perform identification and categorization in a single step with additional end-to-end optimization. YOLO, in particular, offers the finest all-around detection performance in speed, accuracy, and precision. Radar, optical detection, and acoustic sensors are the most often used technologies for detecting UAVs. \textit{Hoffman et al.} investigated the radar detection of UAVs based on separating the Doppler signatures of distinct UAVs \cite{r19}. \textit{Mahnoor et al.} \cite{4} demonstrated that visual images combined with deep learning algorithms solved the UAV detection problem with good precision. According to \textit{Zeeshan et al.}, \cite{1} an acoustic array, unlike radar detection and optical detection approaches, does not rely on the size of the viewed item for detection but on the rotors' sound, and its prerequisite is a large sound dataset. \textit{Maham et al}. \cite{3} performed UAV detection with IR images, but for real-time, the direction detection system could face a multi-class problem. That's why in this paper, we perform multi-class and multi-target drone detection in challenging weather conditions based on infrared (IR) images utilizing an improved YOLOv5p2 model with the main contributions listed below. 
\vspace{-3mm}
\begin{figure*}
  \begin{center}
  \includegraphics[width=4.5in]
  {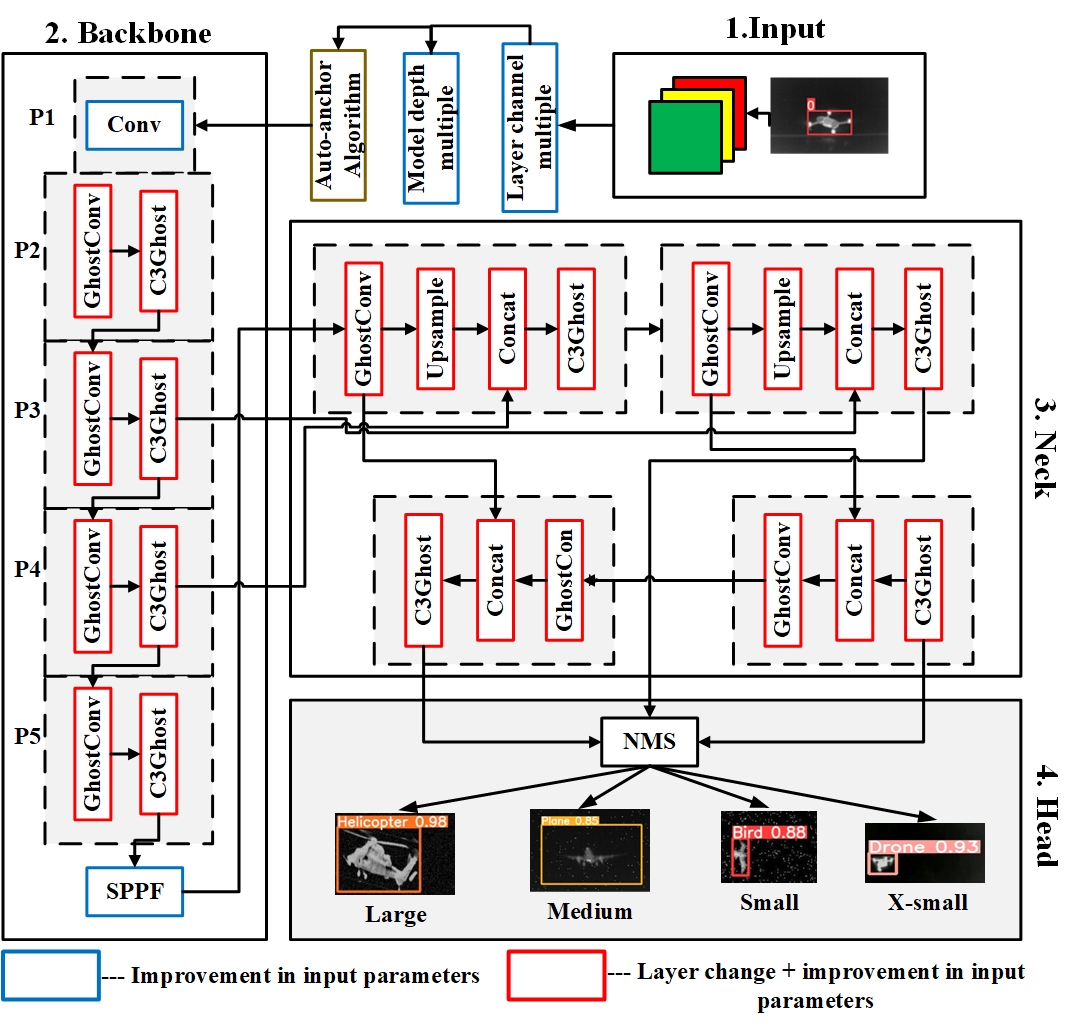}
  \caption{Network topology of the proposed Ghost Auto Anchor Network (GAANet).}
  \label{f1}
  \end{center}
  \vspace{-5mm}
\end{figure*}
\begin{itemize}
  \item Customized dataset \cite{8} is created for multi-class IR-images detection and classification with challenging weather conditions and multi-size targets with varying altitudes.
   \item Improved the baseline YOLOv5 model by introducing an auto-anchor algorithm because it avoids the requirement to scan the input image using a sliding window that computes a prediction at each possible spot.
   \item The night-vision IR images have very small or tiny objects, and we introduced an extra-small anchor (P2) in the model's head. 
   \item Upgraded the baseline's standard Conv and C3 modules with GhostConv and Ghost, respectively \cite{6}. The integration of these modules in the baseline performs optimization at each layer because the ghost module only selects the best and non-repetitive feature maps.
   \item Integration of AdamW \cite{7} optimizer improved the weight decay w.r.t loss function during training as it excellently decouples weight decay from the gradient update step. We named this improved model Ghost auto anchor Network (GAANet).
\end{itemize}

\section{Literature Review}
Drone detection with CNN was envisaged to overcome these limitations because deep neural networks have excellent feature extraction capabilities. Authors in \cite{r8} performed drone identification based on U-Net, a segmented network, to extract regions of interest (ROI), followed by ResNet to categorize the objects in the ROI. Recurrent correlation network (RCN) was used with stationary cameras to improve drone detection in 4K videos\cite{r9}, and correlation filtering to extract the motion features of tiny flying objects. Target identification is skewed when the foreground target is retrieved improperly or incompletely during the background-modeling phase. For the accurate location of the targets in the extracted areas, a two-stage model was outlined in \cite{r10} that used background subtraction to find prospective targets with CaffeNet for target identification. Object detection with computer vision improved the categorization and localization of targets. Such as YOLOv5, Faster-RCNN [4], and CenterNet \cite{r11}. The authors of \cite{r12} created an artifactual data set of drones and birds by removing the target's background and merging them with other images. This dataset was then classified using YOLOv2. Darknet was used as the YOLOv4 backbone for UAV vs. bird detection with an accuracy of 98.3\% \cite{r13}. One-stage detection-based YOLOv5 has excellent detection accuracy and rapid inference time, but its accuracy rapidly drops when dealing with smaller objects in high-resolution photos. The authors in \cite{r14} improved YOLOv5s for multi-rotor UAVs identification. They swapped the YOLOv5s core with Efficientlite, which simplified the model by removing irrelevant layers. The researchers in \cite{r15} fine-tuned the YOLOv5 model with visual images to achieve 95.2\% accuracy and compared the performance with benchmarks like YOLOv3, YOLOv4, and maskRCNN.  In \cite{r16}, authors developed an autonomous drone detection system with sensor fusion of thermal and infrared cameras, which resulted in fewer false positives. In \cite{r17}, the researchers used ResNet as a feature extractor with multi-cascaded auto-encoders for eliminating rain patterns in the UAV images. This method achieved average recognition accuracy of 82\% at 24 frames per second.  An efficient two-stage approach was proposed in \cite{r18} to overcome the problems of high-resolution and small-size UAVs with fixed cameras. The effectiveness of high-resolution images was enhanced by excluding multiple background regions and targeting the candidate regions by SAG-YOLOv5, which had a Ghost module and attention mechanism (SimAM). During an extensive literature review, we deduced that YOLOv5 could not be implemented directly for multi-class UAV detection because of the drone's small size. Also, it faces an exact allocation of bounding boxes with variable-sized targets in a dataset. In this research, we improved YOLOv5 for rapid detection and developed the lightweight model \textit{Ghost auto anchor net (GAANet)} for multi-scale tiny object detection with higher accuracy.

\section{Ghost Auto Anchor Network (GAANet)}
To effectively process the datasets, the CNN include a significant number of parameters, and to minimize them, CNN uses filters\cite{1}\cite{2}. Object detection \cite{3} necessitates many feature maps, each of which has hundreds of channels, making the model bloated and enormous \cite{4}. Therefore, model compression is necessary for rapid deployment on embedding devices with fewer parameters \cite{5}. \textit{Han et al.} presented a novel approach called GhostNet \cite{6} to produce feature maps with fewer operations with reduced duplicate parameters and resource consumption, which allowed the deployment of the trained models on embedded devices quite conveniently. The generation of repetitive, redundant output feature maps with a large number of FLOPs and parameters is the \textbf{ghost} of a handful of intrinsic feature maps with some cheap transformations. These intrinsic feature maps are often smaller and produced by ordinary convolution filters. Here, $m$ intrinsic feature maps $Y^{\prime} \in R^{h^{\prime}\cdot w^{\prime}\cdot m}$ are generated using a standard convolution:
\begin{equation}
Y^{\prime}=X * f^{\prime}
\end{equation}
where $X$ is the input data, $f^{\prime}$ is applied filters and $Y^{\prime}$ is the output feature map, $R$ is required resources, $h^{\prime}$ and $w^{\prime}$ are the height and width of the input data. To further obtain the desired n feature maps, a series of cheap linear operations \textit{ghost based operations} on each intrinsic feature in $Y^{\prime}$ to generate $s$ ghost features according to the following function:
\begin{equation}
y_{i j}=\Phi_{i, j}\left(y_i^{\prime}\right), \quad \forall i=1, \ldots, m, \quad j=1, \ldots, s.
\end{equation}
where $y_i^{\prime}$ is intrinsic feature map in $Y^{\prime}$, $\Phi_{i, j}$ is the linear operation for generating the ghost feature map $y_{i j}$. We effectively used the GhostConv and C3Ghost modules for performing optimized convolutions with the extraction of the most pertinent and unrepeated feature maps with no duplicate gradient information by maintaining accuracy with reduced complexity \cite{6}. The complete network topology of the proposed GAANet is shown in Fig. \ref{f1}, where the size of the input image is set to 265$\times$256 because lower-resolution images increase the generalizability of the GAANet and make it less prone to overfitting, with a focus on important high-level features. The model's depth and channel multipliers are set to 0.25 and 0.5, respectively. These values are chosen so that the model has the best functionality of ghost modules in a lightweight package. The auto-anchor approach is used to apply a K-means function to the modified dataset labels. Then K-means centroids are used as the beginning conditions for a genetic evolution method. 

Here, 1000 generations are investigated before the final calculation of the proposed anchors with CIoU (complete intersection over union) loss and best potential recall as the fitness function. These proposed anchors achieved fitness value of 81.08\%. Now the dataset is passed to the first block of the GAANt backbone, block P1, which extracts extra(x)-small-sized feature maps with an input channel size ($in_{cs}$) of 128, an output channel size ($o_{cs}$) 6, kernel size ($k_s$) of 2, and a stride ($sd$) 2. Block P2 extracts x-small-sized feature maps with $in_{cs}$ of 256, P3 extracts small-sized feature maps with $in_{cs}$ of 512, P4 extracts medium-sized feature maps with $in_{cs}$ of 768, and P5 extracts large-sized feature maps with $in_{cs}$ of 1024. The $o_{cs}$ and $k_s$ of P2, P3, P4, and P5 blocks are fixed at 3 and 2, respectively. The last element of the backbone is SPPF, which does aggregate to eliminate clipping or distortion and disregards the network's fixed-size limitation for the GAANet. The GAANet block determines the locations of the bounding boxes ($x,y$, height, and breadth), scores, and object classes to produce an output image with a bounding box around the identified item and its confidence score.
\section{Model Evaluation}
 \subsection{Dataset and Model Training}
 Here, we proposed GAANet to successfully extract features from collected IR data in low- or no-light conditions at night. To train the proposed GAANet architecture, we gathered around \textbf{5105 IR images} of birds, drones, planes, and helicopters from the publicly available open-source datasets provided on Roboflow. These images also contain different-sized (x-small, small, medium, and large) targets, which make GAANet sensitive to multi-class, multi-size, and multi-type images. In a dataset of 4792 images \cite{8}, 4.6k images are for model training (95\%) and 240 (5\%) for model validation. All experiments are separately run on the \textit{Google Colab} environment with an NVIDIA Tesla T4 GPU having a low learning rate of 0.001, where GAANet and GhostNet-YOLOv5 \cite{l2} has a  batch size of 256 and 512, respectively. The epochs for both models are set to 500 for both models where GhostNet-YOLOv5 \cite{l2} stopped training at 300 by using early stopping as the model performance stopped improving while GAANet stopped training at 457 epochs. 
\begin{table*}
\centering
\caption{Detailed comparison of evaluation metrics}
\begin{tabular}{|p{0.5in}|p{0.5in}|p{0.5in}|p{0.5in}|p{0.5in}|p{0.5in}|p{0.5in}|} \hline 
\textbf{Class} & \multicolumn{3}{|p{1.6in}|}{\ \ \ \ \ \ \ \ \ \ \ \ \ \ \ \ \textbf{GAANet}} & \multicolumn{3}{|p{1.6in}|}{\textbf{\ \ \ \ \ \ \ \ \ GhostNet-YOLOv5\cite{l2}}} \\ \hline 
 & \textbf{Precision} & \textbf{Recall} & \textbf{mAP@0.5} & \textbf{Precision} & \textbf{Recall} & \textbf{mAP@0.5} 
 \\ \hline 
\textbf{Bird} & 97.7 & 95.4 & \textbf{98.6} &97.8 &98.9 &98.5  \\ \hline 
\textbf{Drones} & 90.4 & 98.3 & 97.4 &87 &98.3 &98.4  \\ \hline 
\textbf{Helicopter} & \textbf{99} & 68.8 & 95.9 &99 &57.6 &86.4 \\ \hline 
Plane & 96.7 & \textbf{98.3} & 98.4  &94.3 &96.7 &96.9 \\ \hline
\textbf{Overall} & \textbf{96.2} & \textbf{90.2} & \textbf{97.6} &94.8 &87.9 &95.1 \\ \hline 
\end{tabular}
\label{t4}
\vspace{-3mm}
\end{table*}
\subsection{Evaluation and comparison of trained models}
The detailed evaluation of both trained models GAANet and GhostNet-YOLOv5\cite{l2} is performed by the comparison of true positive (TP), true negative (TN), false negative (FN), false positive (FP), mAP, precision, and recall values. The GAANet model achieved the highest TP value of 1.00 for drones and planes and lowest TP of 0.72 for helicopters. GAANet has the highest FN for helicopters and planes at 0.12 and 0.46, respectively. However, GhostNet-YOLOv5 achieved the highest TP of 1.00 for planes and the lowest TP of 0.47 for helicopters. These stats proved the improved and accurate detection ability of GAANet for planes (1.00 TP), drones (1.00 TP), and birds (0.99 TP), with only helicopters having a TP less than 90\% (0.72).  The addition of ghost-based convolutions and C3 led to the extraction of the most relevant feature maps with smaller channel sizes than the baseline model, resulting in reduced model size, layers, parameters, and GFLOPs. The proposed GAANet trained with 395 layers took 3.210 hours to train on a completely customized dataset with a weight size of 14.1 MB. 

\begin{figure}[h]
     \centering
     \begin{subfigure}[b]{0.1\textwidth}
         \centering
\includegraphics[width=\textwidth]{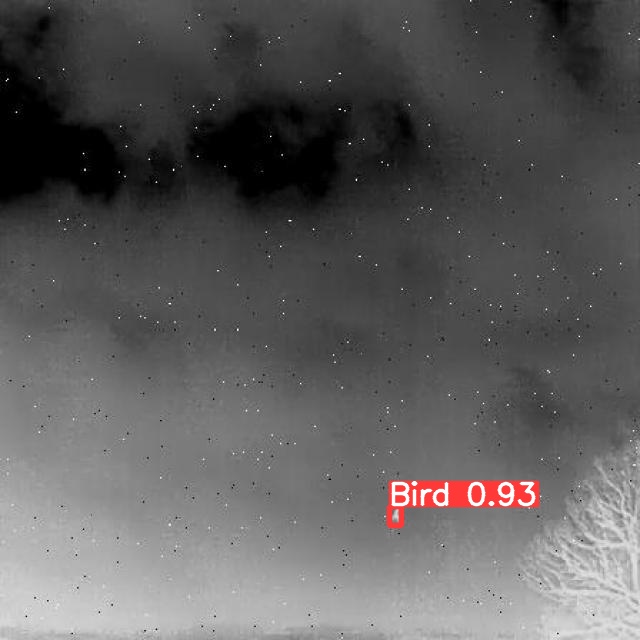}
\includegraphics[width=\textwidth]{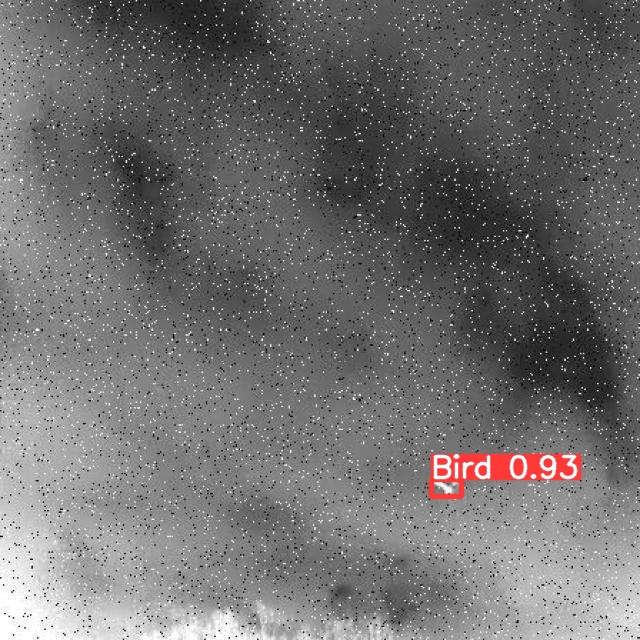}
\includegraphics[width=\textwidth]{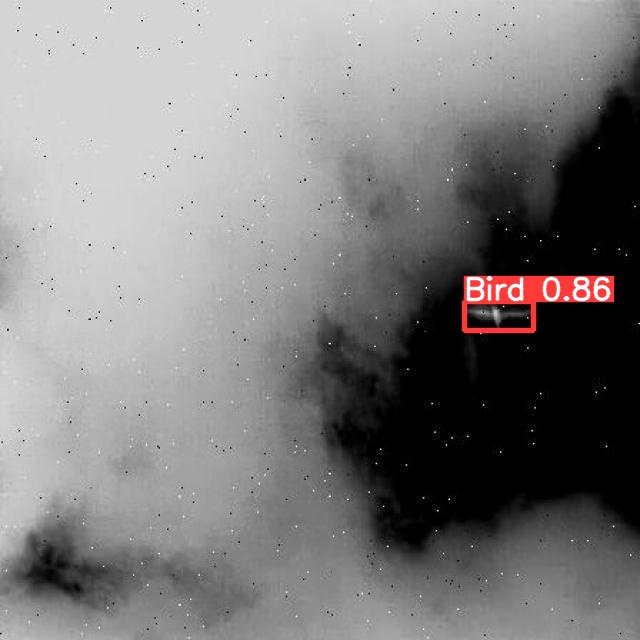}
\includegraphics[width=\textwidth]{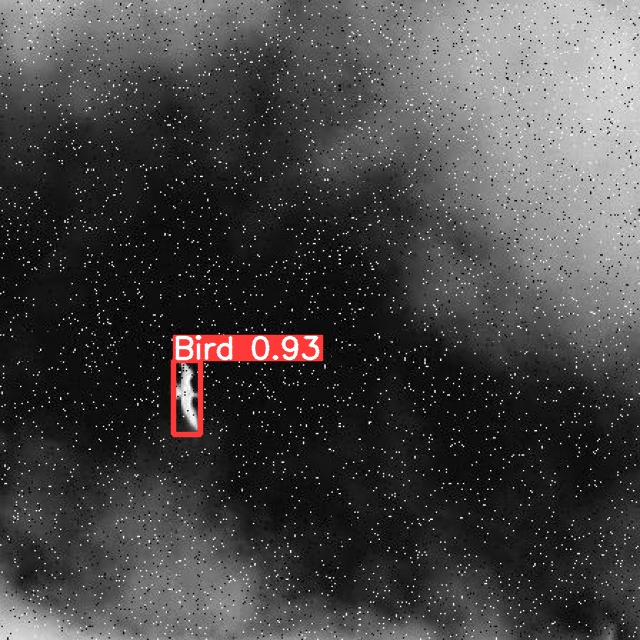}
         \caption{Bird}
         \label{figa}
     \end{subfigure}
     \hfill
     \begin{subfigure}[b]{0.1\textwidth}
         \centering
         \includegraphics[width=\textwidth]{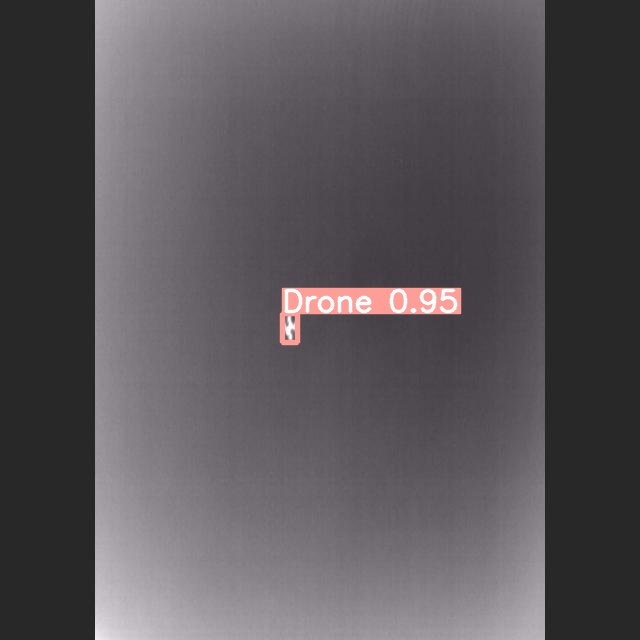}
         \includegraphics[width=\textwidth]{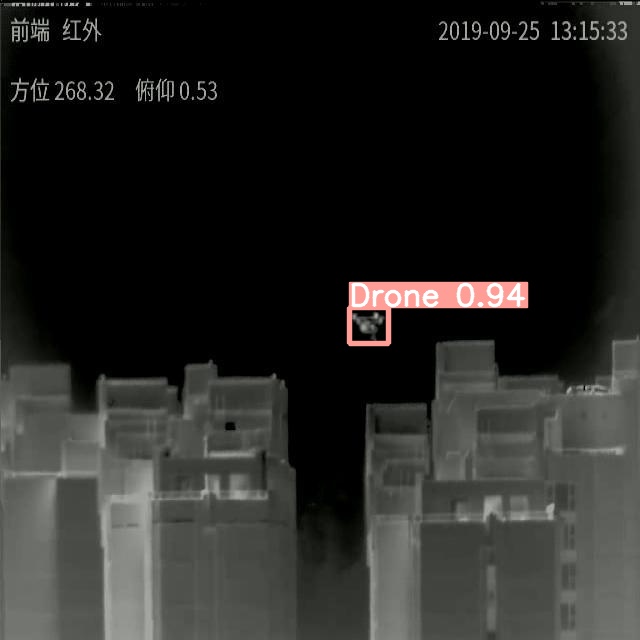}
         \includegraphics[width=\textwidth]{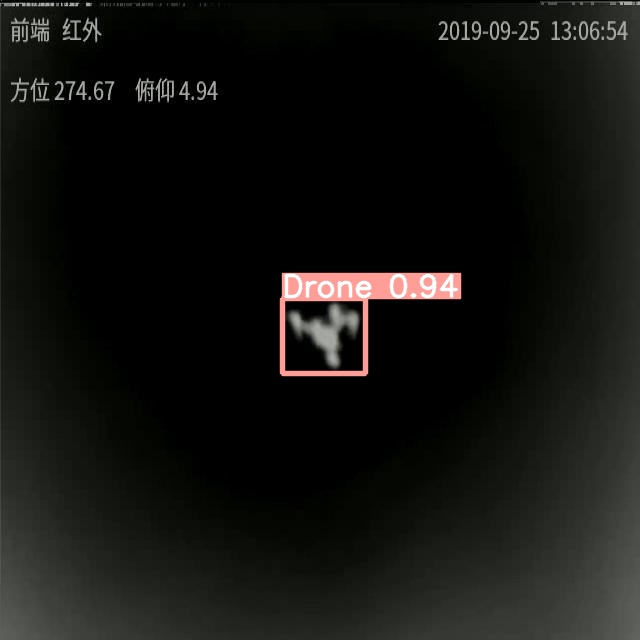}
         \includegraphics[width=\textwidth]{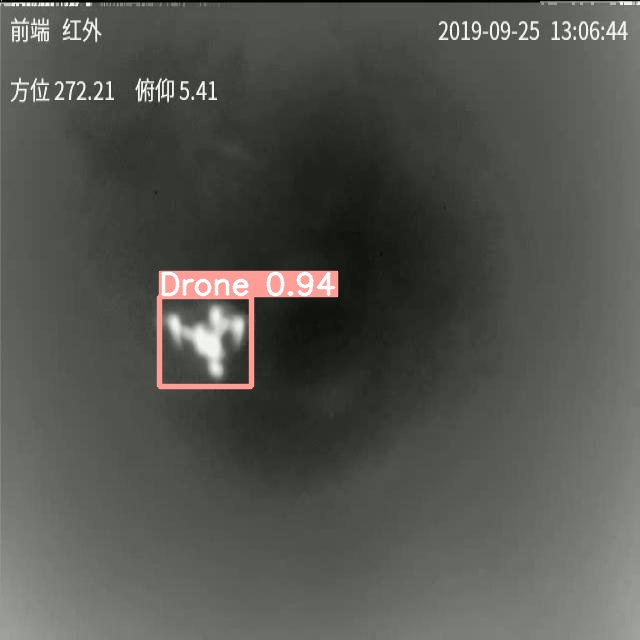}
         \caption{Drone}
         \label{figb}
     \end{subfigure}
     \hfill
     \begin{subfigure}[b]{0.1\textwidth}
         \centering
          \includegraphics[width=\textwidth]{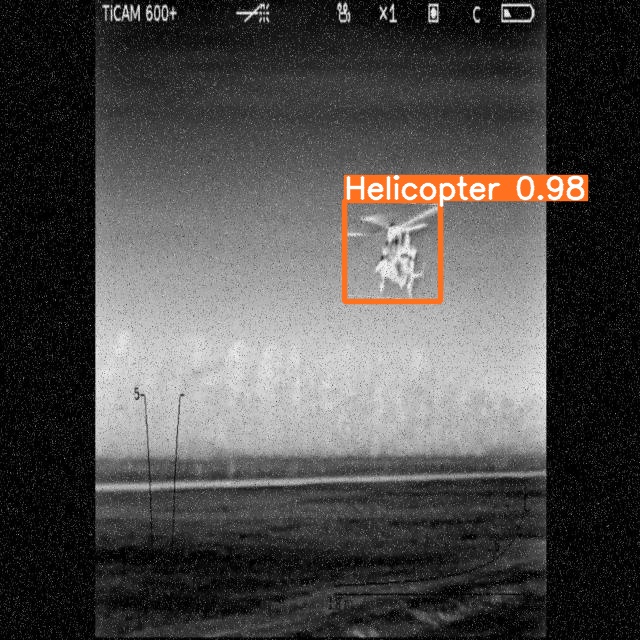}
         \includegraphics[width=\textwidth]{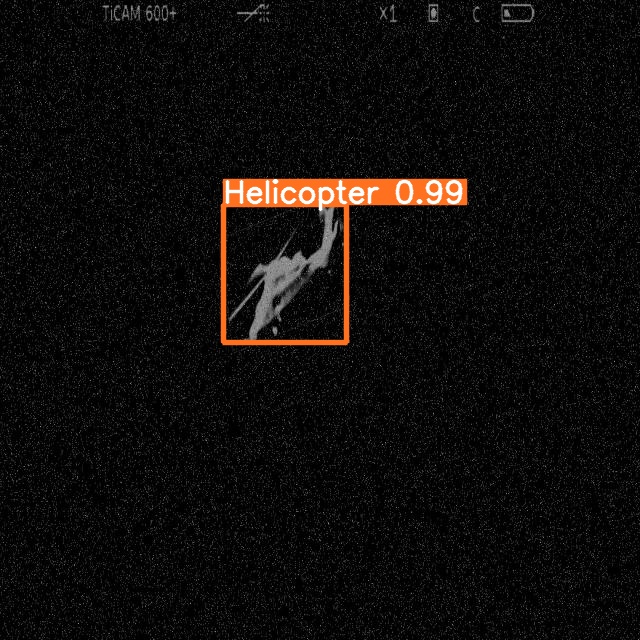}
         \includegraphics[width=\textwidth]{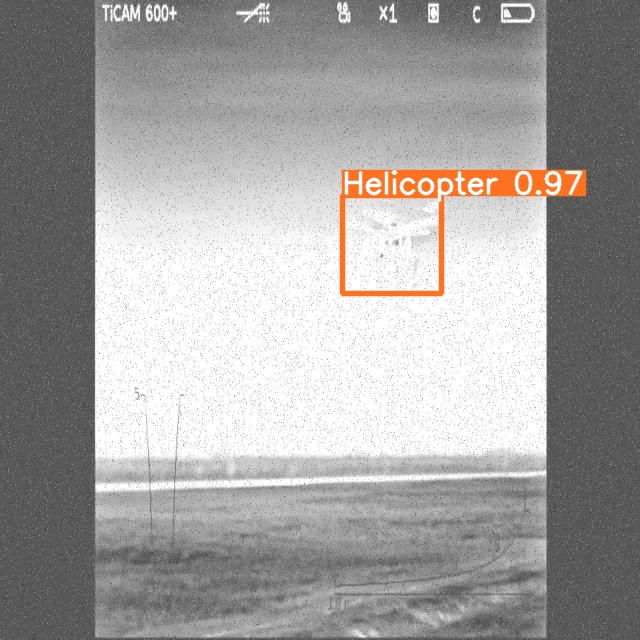}
         \includegraphics[width=\textwidth]{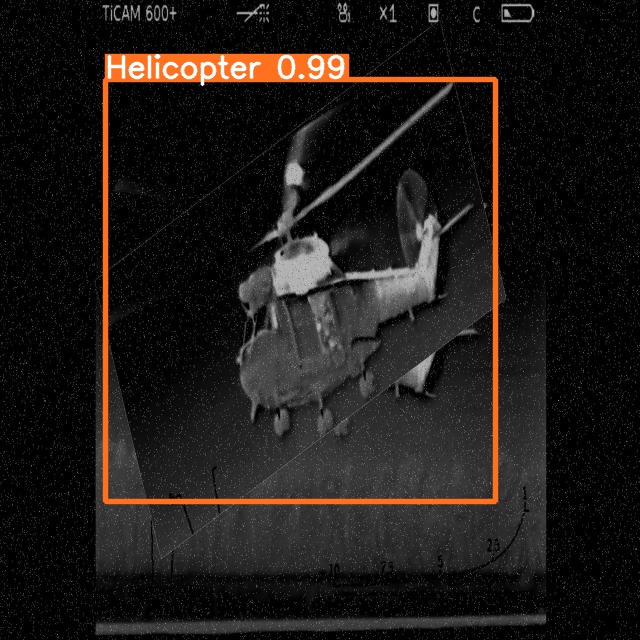}
         \caption{Helicopter}
         \label{figc}
     \end{subfigure}
     \hfill
     \begin{subfigure}[b]{0.1\textwidth}
         \centering
         \includegraphics[width=\textwidth]{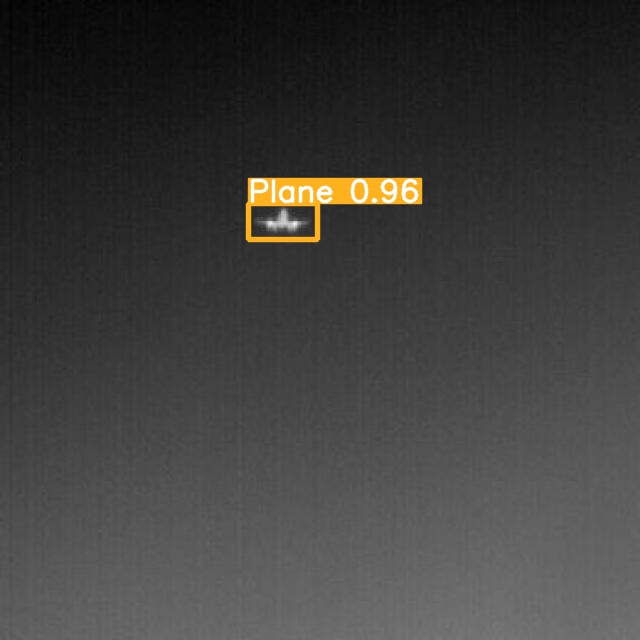}
         \includegraphics[width=\textwidth]{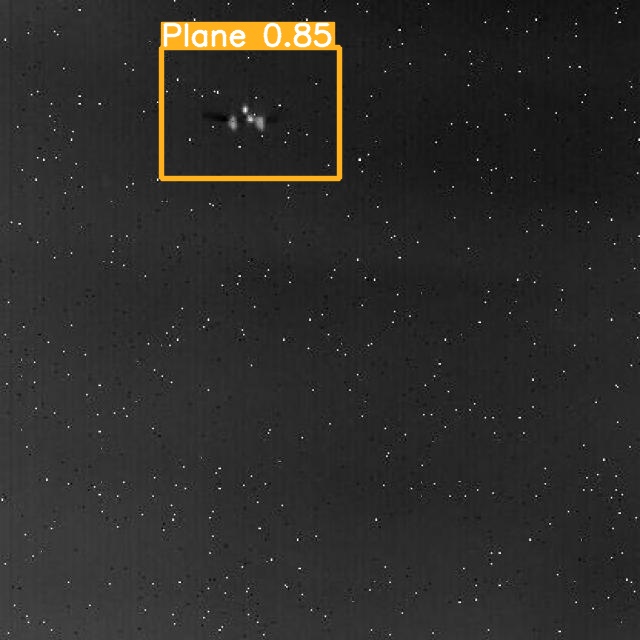}
         \includegraphics[width=\textwidth]{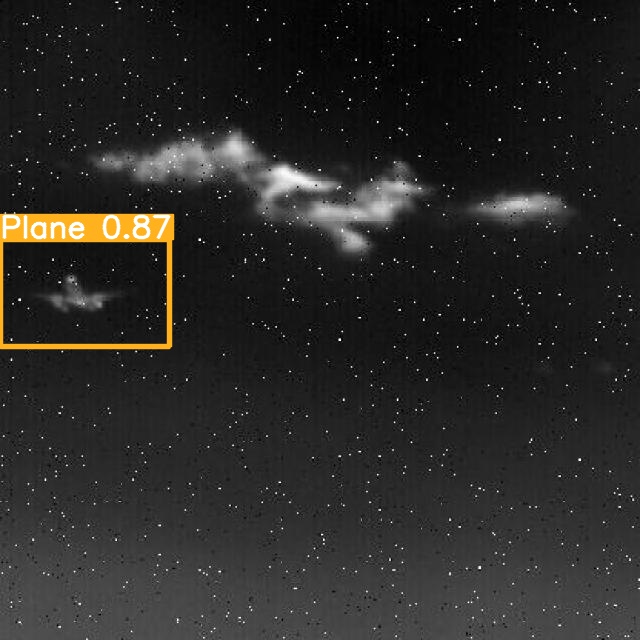}
         \includegraphics[width=\textwidth]{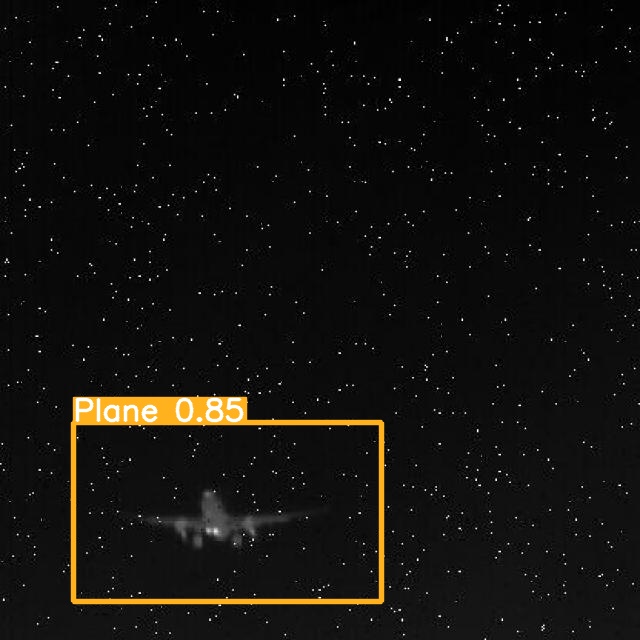}
         \caption{Plane}
         \label{figd}
     \end{subfigure}
     \hfill
     \caption{Detection results of Multi-size IR targets (Top to bottom) x-small, small, medium and large (a-d) GAANet.}
    \label{f2}
    \vspace{-5mm}
\end{figure}

\begin{figure}[h]
     \centering
     \begin{subfigure}[b]{0.1\textwidth}
\includegraphics[width=\textwidth]{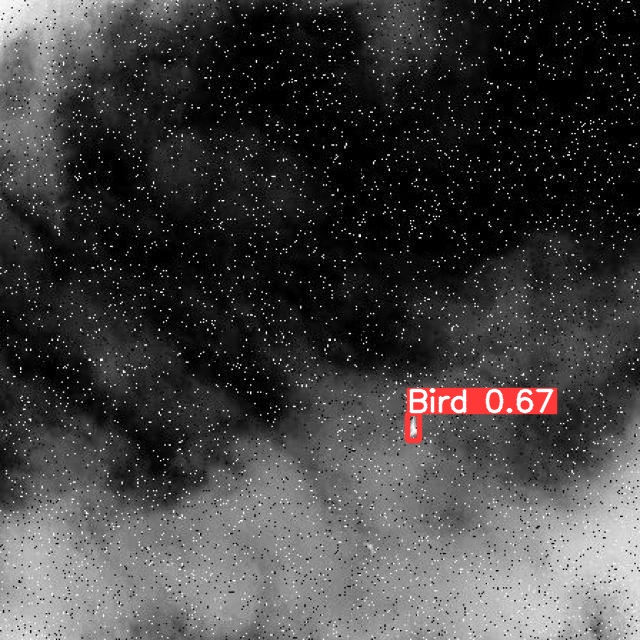}
\includegraphics[width=\textwidth]{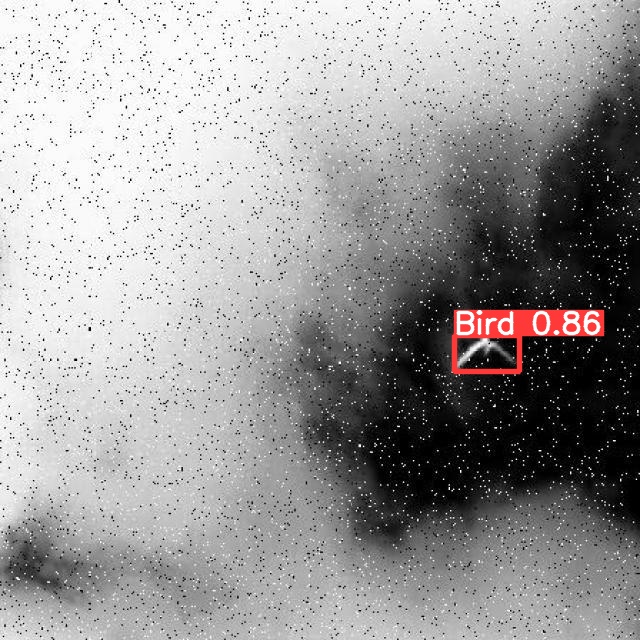}
\includegraphics[width=\textwidth]{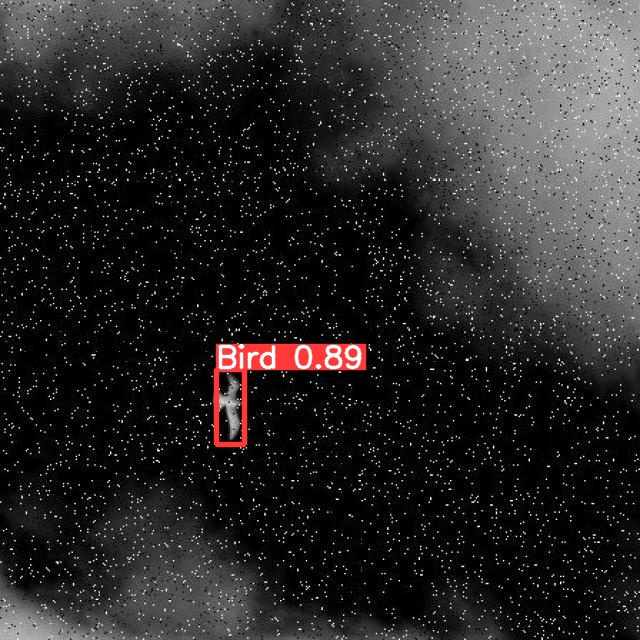}
         \caption{Bird}
         \label{figa}
     \end{subfigure}
     \hfill
     \begin{subfigure}[b]{0.1\textwidth}
         \centering
         \includegraphics[width=\textwidth]{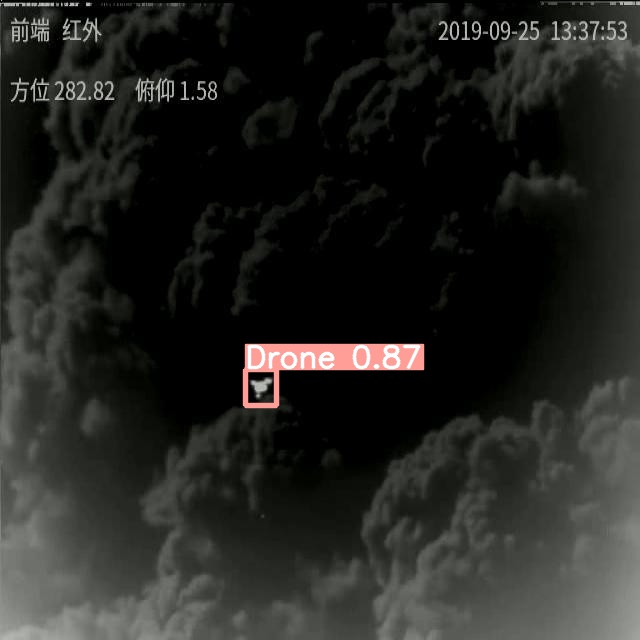}
         \includegraphics[width=\textwidth]{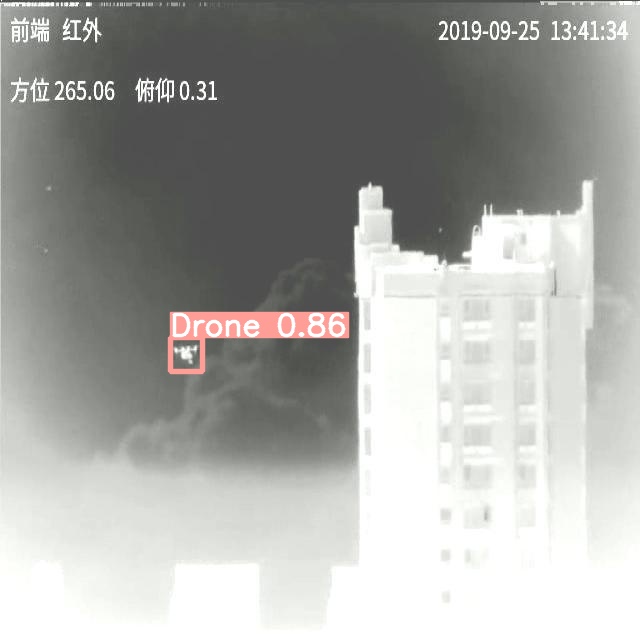}
         \includegraphics[width=\textwidth]{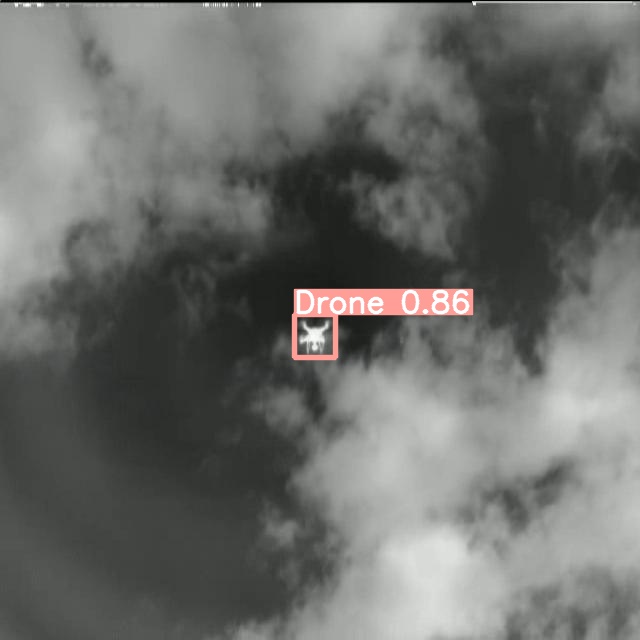}
         \caption{Drone}
         \label{figb}
     \end{subfigure}
     \hfill
     \begin{subfigure}[b]{0.1\textwidth}
         \centering
         \includegraphics[width=\textwidth]{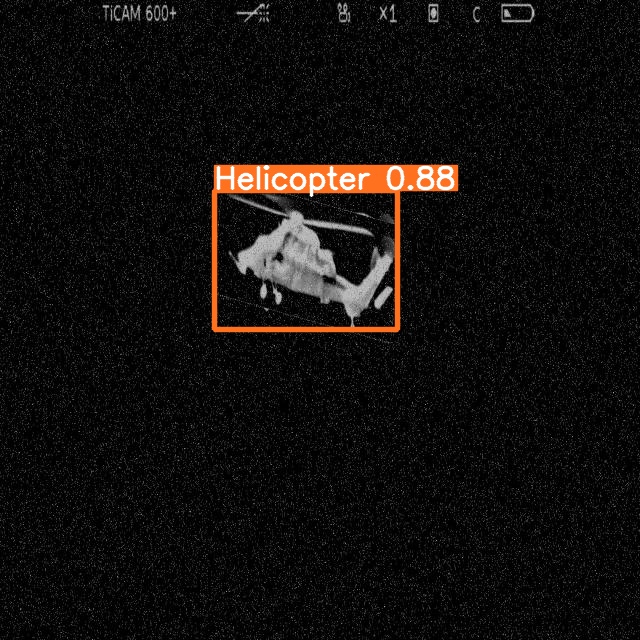}
         \includegraphics[width=\textwidth]{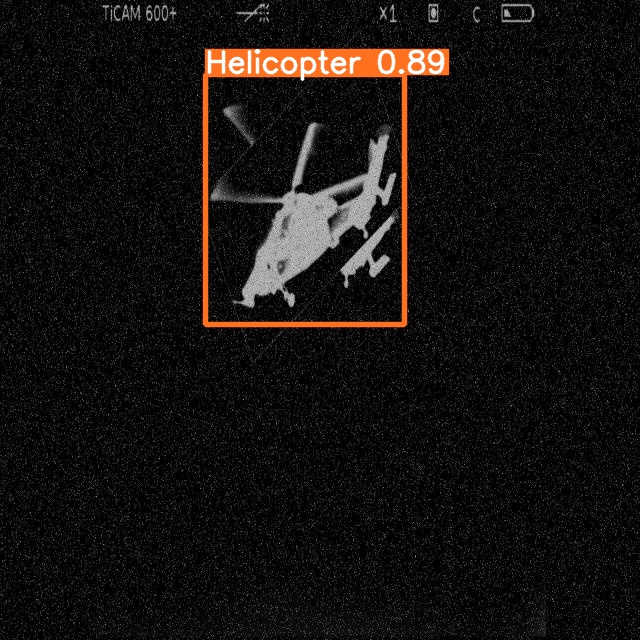}
         \includegraphics[width=\textwidth]{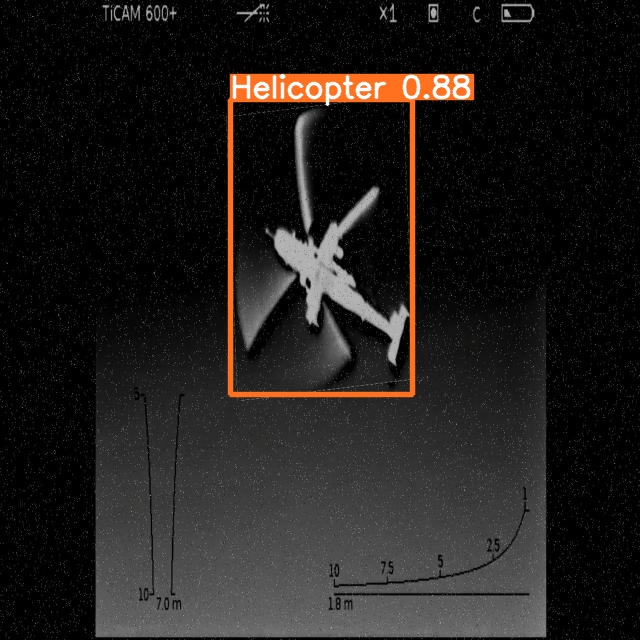}
         \caption{Helicopter}
         \label{figc}
     \end{subfigure}
     \hfill
     \begin{subfigure}[b]{0.1\textwidth}
         \centering
         \includegraphics[width=\textwidth]{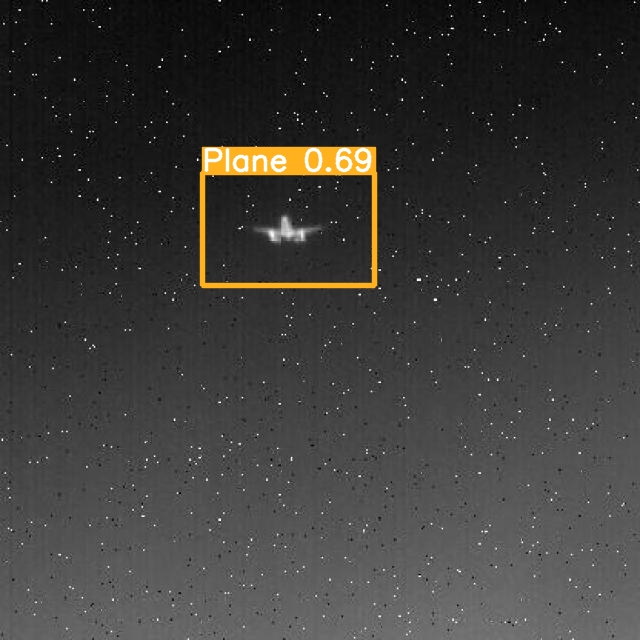}
         \includegraphics[width=\textwidth]{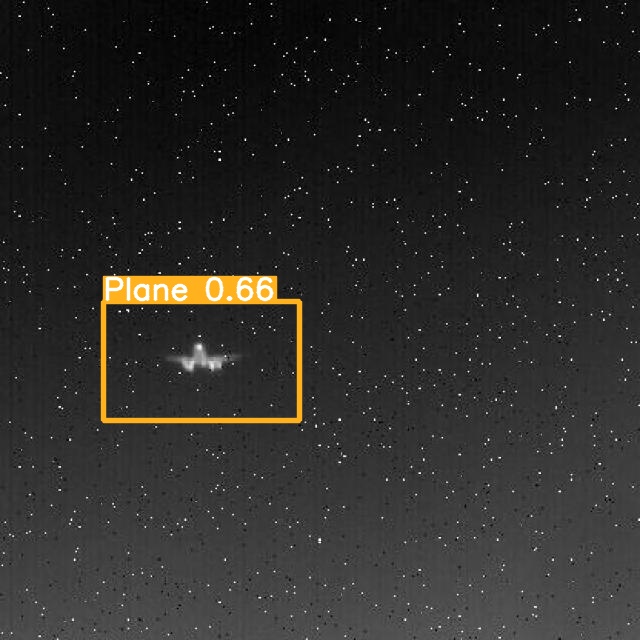}
         \includegraphics[width=\textwidth]{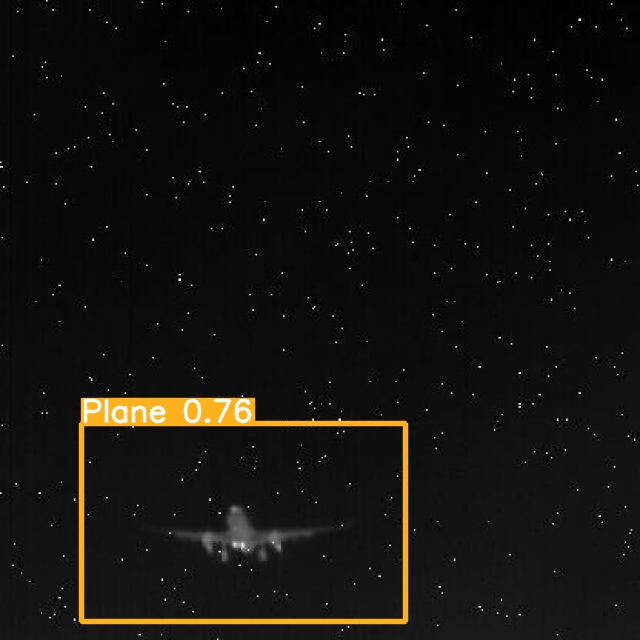}
         \caption{Plane}
         \label{figd}
     \end{subfigure}
     \hfill
     \caption{Detection results of Multi-size IR targets (Top to bottom) small, medium and large (a-d) GhostNet-YOLOv5 \cite{l2}.}
     \vspace{-3mm}
    \label{f3}
\end{figure}

The layer size is reduced due to the smaller value of channel and depth of the GAANet model compared to the baseline model. GAANet has the highest object precision of 99\% for helicopters of varied sizes and altitudes, whereas planes have the best recall value of 98.3\% and birds with the highest $mAP@0.5$ of 98.6\%. GAANet has the lowest precision value of 90.4\% for drones which is 21.5\% more than GhostNet-YOLOv5, as it achieved 68.9\% detection precision for drones. Similarly, the 18.4\% higher recall is achieved for helicopters compared to GhostNet-YOLOv5.  Therefore, the overall precision of GAANet is increased by 1.4\%, recall by 2.3\%, and $mAP@50$ by 2.5 \% compared to GhostNet-YOLOv5. The average inference time achieved by GAANet over a batch of multiple images is 13.7 ms, which is 2.7 ms less than GhostNet-YOLOv5 of 16.4 ms.  Fig. \ref{f2} and Fig. \ref{f3} show the detection results of GAANet and GhostNet-YOLOv5 tested with unseen and unknown IR images. For GAANet bird detection, the x-small and small IR images have the highest detection accuracy of 0.93. For drone IR images, GAANet achieved 0.94 accuracies for x-small images, while small, medium, and large IR drone images have 0.93 detection accuracy. GAANet achieved 0.99 accuracies on small and large helicopter IR images. For plane detection, GAANet attained 0.96 accuracies for x-small plane IR images, while small, medium, and large IR plane images have 0.85, 0.87, and 0.85 accuracies, respectively. From these results, we can infer that GAANet improved performance on all sized target IR images, but it attained the best accuracy for the x-small bird, drone, and plane IR images. GhostNet-YOLOv5 had the best testing accuracy for large bird IR images (0.89), small drone IR images (0.87), medium helicopter IR images (0.89), and large plane IR images (0.76).

\subsection{Comparison with state-of-the art}
The authors in \cite{3} performed classification with the TFNet model for drone vs. bird IR images. In comparison with GAANet, TFNet achieved lower precision, recall, and mAP of around 0.5\%, 12.8\%, and 13.6\%, respectively. The MFNet-M model \cite{4} performed UAV vs. bird detection using visual images. Their proposed MFNet-M model achieved 94\% recall with 95.9\% mAP on drone images, while GAANet has 98.3\% recall with 97.4\% mAP on drone IR images which are 4.3\% and 1.5\% high, respectively. The authors in \cite{5} performed multi-type UAV classification with YOLOv7 on visual images. Single rotor UAV images achieved 88.7\% recall and 94\% mAP, which are 4\% more, 9.6\%, and 2.5\% less than GAANet, respectively. In \cite{l1}, the authors improved the baseline YOLOv4 backbone with GhostNet and achieved 0.9\% less precision than GAANet. Similarly, the authors used Ghost convolution in the YOLOv5 baseline model and achieved 19.89\% less precision than GAANet \cite{l2}. YOLOv5x-ALL-GHOST added GhostNet in both head and backbone and got 0.6\% less $map@0.5$ than GAANet \cite{l3}. GhostNet feature extraction networking was embedded in YOLOv3 \cite{l4} that achieved 8.3\% less $map@0.5$ than GAANet. 
\begin{table}[h]
\caption{Comparison with the state-of-the-art schemes for UAV detection.}
\centering
    \begin{tabular}{p{2.5cm} p{1cm} p{1cm} p{1cm} p{1cm} }
    \hline
    \textbf{Model} &\textbf{Precision (\%)} &\textbf{Recall (\%)} &\textbf{Parameters (million)} &\textbf{Weight (MB)}
    \\ \hline
    \textbf{MFNet-M}	
    &96.8 &90.4 &5.2 &75.3	
    \\ \hline
    \textbf{YOLOv4-GhostNet \cite{l1}} &95.32 &86.54 &39.70 &150
    \\ \hline
    \textbf{GhostNet-YOLOv5 \cite{l2}} &76.31 &88.42 &5.9 &10
    \\ \hline
    \textbf{YOLOv5x-ALL-GHOST \cite{l3}} &N/G &N/G &25.09 &48.7
    \\ \hline
    \textbf{YOLO-G \cite{l4}} &88.9 &86.3 &N/G &42.7
    \\ \hline
    \textbf{Proposed GAANet} &96.2 &90.2 &6.8 &14.1
    \\ \hline
    \label{t6}
    \end{tabular}
    \vspace{-8mm}
\end{table}
\section{Conclusion}
In this paper, we proposed an improved and optimized deep-learning model for extra small-sized flying object detection, specifically UAVs, using IR images during night surveillance. The proposed GAANet used ghost convolution, ghost C3, and a downsampled input channel size to extract the most prominent and non-repeated features. Detailed experimentation was performed on a customized multi-class dataset containing drones, planes, helicopters, and birds. The results showed a low misclassification rate, which confirms the effectiveness of the proposed model for real-time night vision IR images. The proposed object detection approach outperformed the other current state-of-the-art technologies by a significant margin.
\section{Acknowledgment}
This work was supported by the Higher Education Commission (HEC) Pakistan under the NRPU 2021 Grant 15687 and also by Engineering and Physical Sciences Research Council (EPSRC), U.K., under Grant EP/W004348/1.

\bibliographystyle{ieeetr}
\bibliography{GAANet.bib}

\end{document}